\documentclass[runningheads]{llncs}
\usepackage[T1]{fontenc}
\usepackage{graphicx}
\usepackage{amsmath}
\usepackage{cleveref}
\usepackage{multirow}
\usepackage{array}
\usepackage{booktabs}

\newcolumntype{C}[1]{>{\centering\arraybackslash}p{#1}}

\begin{document}
%
\title{Unbiased Regression Loss for DETRs}
%
%

\author{Edric\inst{1} \and
Ueta Daisuke\inst{2} \and
Kurokawa Yukimasa\inst{2} \and
Karlekar Jayashree\inst{1} \and
Sugiri Pranata\inst{1}
}

\authorrunning{Edric et al.}

\institute{Panasonic R\&D Center Singapore \and 
Panasonic Connect Co., Ltd. R\&D Division
}
\maketitle              
\begin{abstract}
In this paper, we introduce a novel unbiased regression loss for DETR-based detectors. The conventional $L_{1}$ regression loss tends to bias towards larger boxes, as they disproportionately contribute more towards the overall loss compared to smaller boxes. Consequently, the detection performance for small objects suffers. To alleviate this bias, the proposed new unbiased loss, termed Sized $L_{1}$ loss, normalizes the size of all boxes based on their individual width and height. Our experiments demonstrate consistent improvements in both fully-supervised and semi-supervised settings using the MS-COCO benchmark dataset.

\end{abstract}

\section{Introduction}
\label{sec:intro}

\begin{figure}[t]
    \centering
    \includegraphics[width=7.5cm]{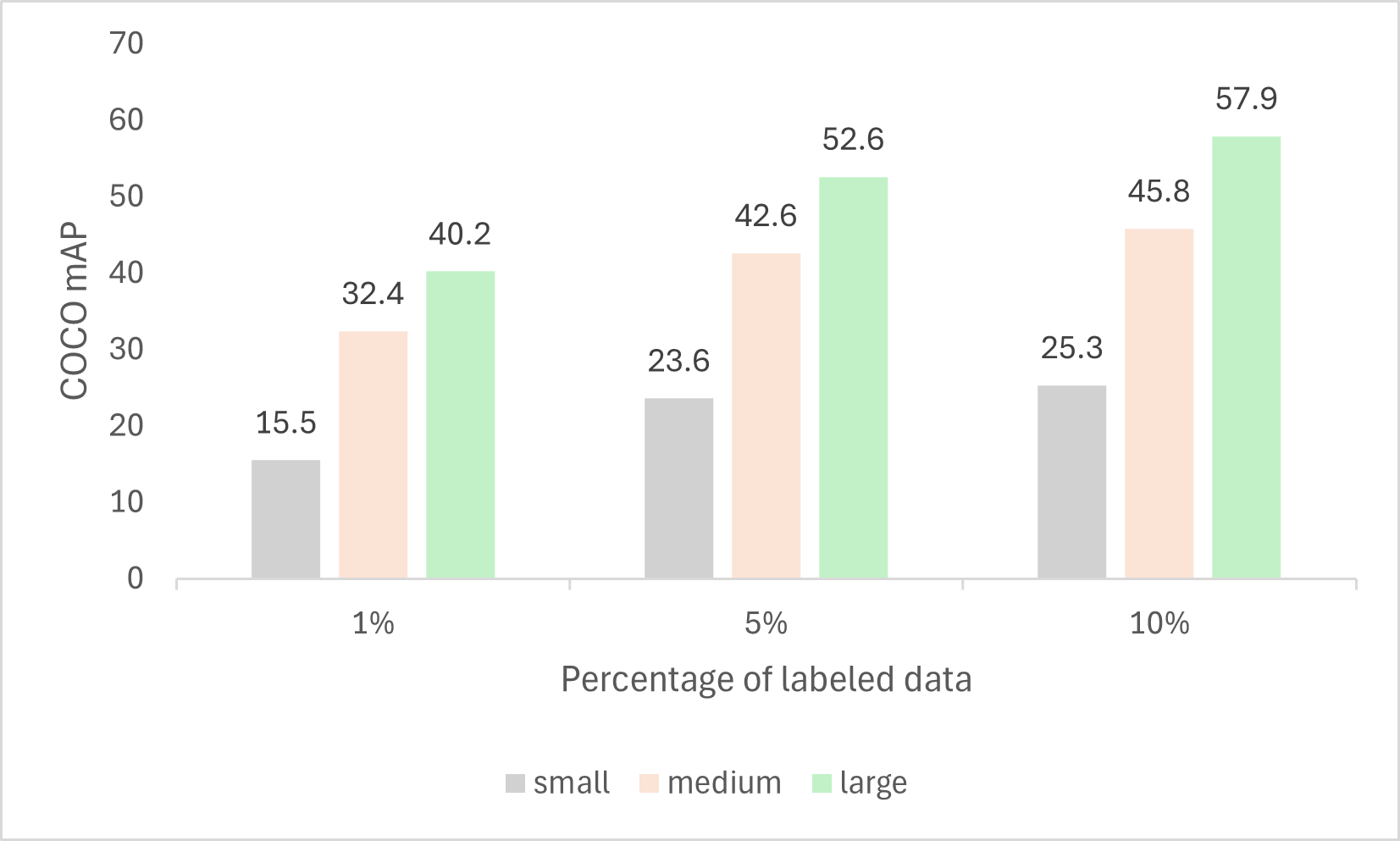}
    \caption{Discrepancy in COCO $mAP$ for small, medium, and large boxes with Semi-DETR}
    \label{Discrepancy in $mAP$}
\end{figure}

Object detection models traditionally necessitate numerous hand-crafted components to facilitate training, including non-max suppression (NMS) \cite{bodla2017soft}, a set of predefined anchor boxes, label matching, etc. 
Conversely, DETR (DEtection TRansformer \cite{carion2020end})-based detectors \cite{zhu2020deformable,li2022dn,liu2022dab,meng2021conditional,zhang2022dino} largely eliminate the need for such components, while maintaining or even enhancing the detection capabilities compared to traditional detectors.
Notably, both traditional and DETR-based detectors exhibit significant disparities in their detection capability of objects of varying sizes, as highlighted in \cref{Discrepancy in $mAP$}.
This issue is partly due to the inherent bias in the loss function, where identical deviations in smaller boxes contribute less to the overall loss, i.e., the loss is scale-variant. 
To address this, traditional detectors employ scale-invariant losses in the regression head or normalize the boxes with respect to their anchor boxes. DETR-based detectors utilize a regression loss comprising $L_{1}$ loss and $IoU$ loss \cite{zhou2019iou,rezatofighi2019generalized,zheng2020distance}. Although the $IoU$ loss is scale-invariant, the $L_{1}$ loss is not.
Thus, we aim to mitigate any bias towards larger boxes by unbiasing the $L_{1}$ regression loss and normalizing each boxes with their respective dimensions --- width and height --- to ensure equal contribution of smaller boxes as that of larger ones. We term our proposed normalized loss as Sized $L_{1}$ loss.

We apply this loss to two different scenarios of DETR-based object detection: \textbf{(1) fully-supervised} and \textbf{(2) semi-supervised}. 

In fully-supervised scenario, we utilize Co-DETR \cite{zong2023detrs} as the baseline and replace the $L_{1}$ loss with Sized $L_{1}$ loss in the main DETR head while maintaining the original losses of the auxiliary heads. 

In semi-supervised scenario, we utilize Semi-DETR \cite{zhang2023semi} as our baseline. In a teacher-student framework, the performance is highly influenced by the student's ability to learn from both supervised and unsupervised data as well as the teacher model's capability to generate accurate pseudo-labels. Any bias learned by the student model is propagated back to the teacher in a negative feedback loop. Therefore, we replace the $L_{1}$ regression loss with Sized $L_{1}$ loss. The loss can be replaced across both training branches, supervised and unsupervised, or only within the supervised branch.

Experimental results show that the proposed Sized $L_{1}$ loss improves the baseline performance in both scenarios, particularly for smaller objects.

\section{Related Work}
\subsection{Scale Invariant Regression Loss}
In most object detectors, $L_{1}$ loss or its variants are commonly used for the bounding box regression. In some anchor-based detectors, the $L_{1}$ loss is calculated after the normalization of both predicted and ground-truth boxes with the respective anchor boxes, causing it to be somewhat scale-invariant. Moreover, complete scale-invariant losses, particularly $IoU$ (Intersection over Union)-based losses, have also been used in traditional detectors. Zhou etal \cite{zhou2019iou} replaces $L_{1}$ loss and simply utilizes the overlap between prediction and label as the loss.
Additionally, \cite{rezatofighi2019generalized} and \cite{zheng2020distance} introduce an extra penalty term to address the issue of non-overlapping boxes. 

Although $IoU$-based loss is used as a component of the regression loss in DETR-based detectors, the scale-sensitive $L_{1}$ loss is still used, often assigned greater weight than its IoU-based counterpart. Therefore, we propose to additionally normalize the $L_{1}$ loss for the overall loss to be scale-invariant, thus improving the model's robustness across varying object sizes.

\subsection{Transformer-based Object Detection}
Following the growing interest in transformer-based models across both language and vision fields, DETR \cite{carion2020end} achieves end-to-end detection with highly competitive performance. Using a transformer encoder-decoder architecture, DETR formulates the detection task as a set prediction problem, enabling efficient end-to-end training without relying on the numerous hand-crafted components typical of traditional detectors. 
However, DETR exhibits slow convergence issue partly due to the transformer's high complexity and the sparse supervision from the object queries without objects.
Many works \cite{zhu2020deformable,li2022dn,liu2022dab,meng2021conditional,zhang2022dino} aim to reduce the complexity, increase the number of supervisions by ways of additional denoising, or introduce a better and more efficient initialization of the decoder queries. Recently, DINO \cite{zhang2022dino} combines the various improvements proposed by previous papers. Co-DETR \cite{zong2023detrs} builds upon DINO \cite{zhang2022dino} and further introduces auxiliary detection heads to increase the number of positive supervision. In this paper, Co-DETR is used as the benchmark model for the fully-supervised task.

\subsection{Semi-Supervised Object Detection}
In the SSOD task, a teacher-student learning framework is commonly used, where a student model learns from the available supervised data, and a teacher model - constituting a running EMA-updated version of the student - and generating pseudo labels from unsupervised data as further supervision for the student model. Many works have explored the SSOD task \cite{liu2021unbiased,zhou2022dense,zhou2021instant,xu2021end,tang2021humble,li2022pseco,sohn2020simple}, mainly with conventional detectors such as RCNN detector family \cite{girshick2014rich,he2017mask,girshick2015fast,ren2015faster} or RetinaNet \cite{lin2017focal}.

A recent work \cite{zhang2023semi} applies the task with DETR-based detectors, presenting a new way of enforcing consistency regularization to account for the set-based nature of the object queries in DETR. This is achieved by interchanging the relevant object queries from the teacher and student and adding a new cross-view consistency loss between the two models.

Any bias can be propagated in a negative feedback loop within the teacher-student learning framework; the student learns the bias from the supervised data, propagating the bias to the teacher through the EMA update, and the teacher back to the student via the generated pseudo-labels. We aim to remove any scale bias caused by the $L_{1}$ regression loss through the proposed Sized $L_{1}$ loss.

\section{Proposed Method}
\subsection{Preliminary}
We aim to address the intrinsic size bias present in the $L_{1}$ regression loss function used in DETR-based detectors. 
We extend our studies to both DETR-based SSOD and fully-supervised settings.
Formally, the SSOD task consists of a set of labeled image $D_{l} = \{x_{l}^{i}, y_{l}^{i} \}_{N_{l}}^{i=1}$ and a set of unlabeled image $D_{u} = \{x_{u}^{i}, y_{u}^{i} \}^{N_{u}}_{i=1}$, both of which are available during training. $N_{l}$ and $N_{u}$ denote the amount of labeled and unlabeled images. For the labeled images $x_{l}$, the annotations $y_{l}$ contain the coordinates and object categories of all bounding boxes. 

\begin{figure}[tb]
  \includegraphics[width=\linewidth]{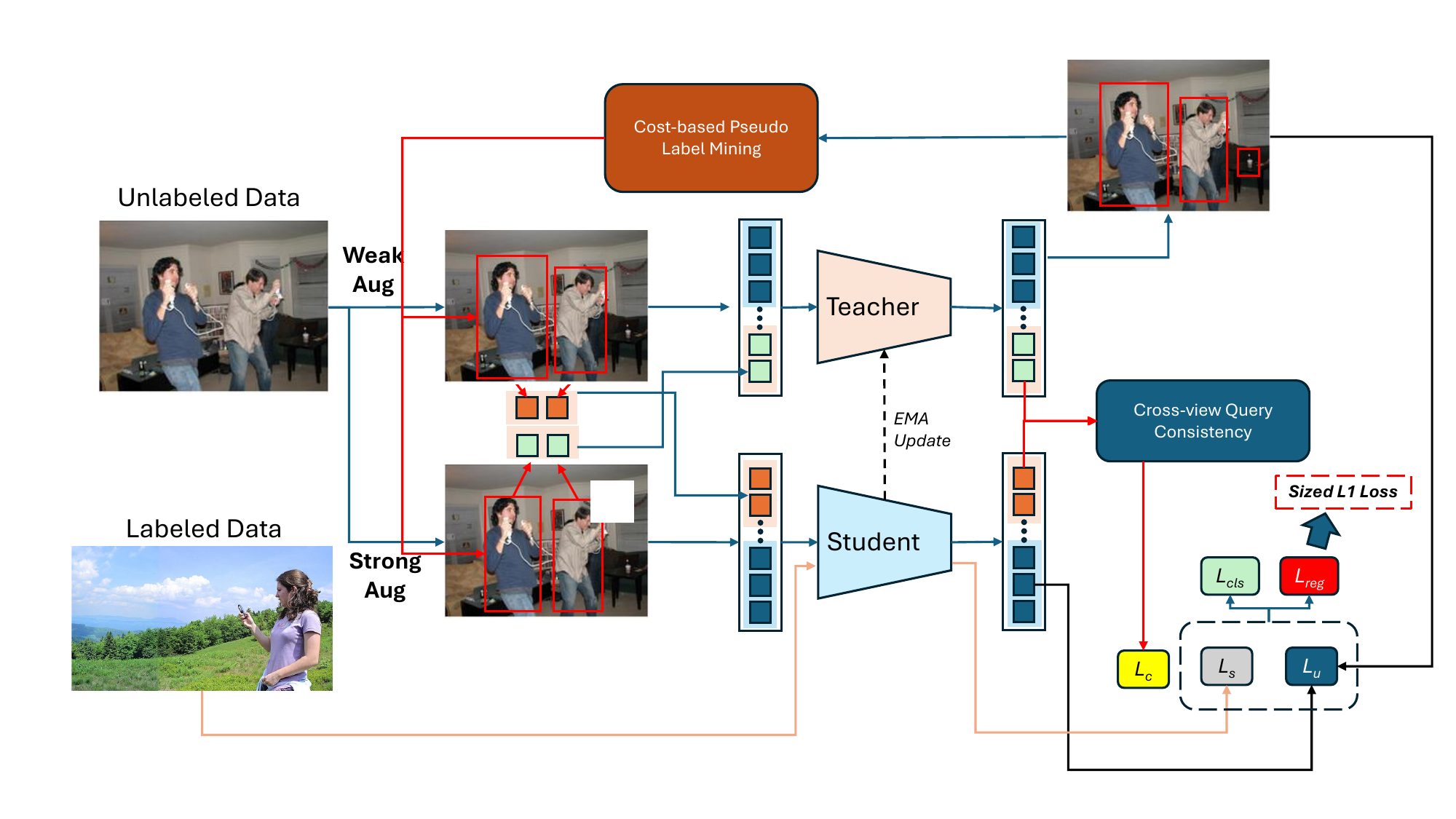}
  \caption{Model overview of Semi-DETR, with the regression component of the supervised and unsupervised losses replaced with Sized $L_{1}$ loss}
  \label{Overview of Sized $L_{1}$ Loss}
\end{figure}

\subsection{Overview}
To validate the effectiveness of the proposed Sized $L_{1}$ loss, we choose the framework established in Semi-DETR \cite{zhang2023semi}, adopting both its model structure and backbone. Specifically, we implement DINO \cite{zhang2022dino} as our model with a ResNet-50 \cite{he2016deep} backbone. Following the teacher-student framework, two differently augmented views of unlabeled images are fed to the teacher and student model - weakly-augmented to the teacher, and strongly-augmented to the student. The student model is trained using labeled data in a standard supervised manner and unlabeled data via pseudo-labels generated by the teacher model. The student model is updated in each iteration by back-propagation whereas the teacher model is an EMA-updated version of the student. Finally, we replace the $L_{1}$ regression loss with the proposed normalized Sized $L_{1}$ loss as shown in \cref{Overview of Sized $L_{1}$ Loss}. Similarly, in the fully-supervised setting, we replace the $L_{1}$ regression loss of the main DETR head with Sized $L_{1}$ loss while maintaining the other losses of the auxiliary heads.

\subsection{Overall Losses}
In Semi-DETR, the training loss integrates a new cross-view consistency loss on top of the supervised $\mathcal{L}_{s}$ and unsupervised $\mathcal{L}_{u}$ losses found in Co-DETR.

\begin{align}
    \mathcal{L} = \mathcal{L}_{s} + \mathcal{L}_{u} \\
    \mathcal{L}_{s} = \mathcal{L}_{s}^{cls} +  \mathcal{L}_{s}^{reg} \\
    \mathcal{L}_{u} = \mathcal{L}_{u}^{cls} +  \mathcal{L}_{u}^{reg} \\
    \mathcal{L}^{reg} = \mathcal{L}_{GIoU} + \mathcal{L}_{L_{1}}
\end{align}

We focus on the regression loss component $\mathcal{L}^{reg}$ of both supervised and unsupervised losses, replacing the $L_{1}$ loss with the Sized $L_{1}$ loss and pairing it with $IoU$ regression loss to produce a more scale-invariant regression loss.

\subsection{Sized $L_{1}$ Loss} 
\begin{figure}[tb]
    \centering
    \includegraphics[width=9cm]{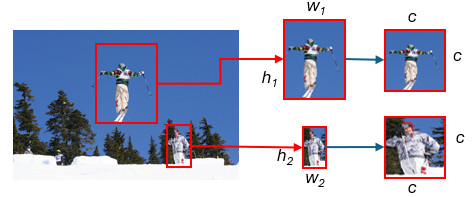}
    \caption{Sized $L_{1}$ loss effectively normalizes all boxes to equal width and height before loss calculation}
    \label{Sized $L_{1}$ loss}
\end{figure}

Conventionally, along with $IoU$ loss, $L_{1}$ loss is used as the regression loss in DETR detectors, particularly for its bounding box regression:

\[
L_{1}(\hat{b}_i, b_i) = \sum_{i,j} w_{i,j} \cdot |\hat{b}_{i,j} - b_{i,j}|
\]

where $\hat{b}_{i,j}$ and $ b_{i,j}$ represent the predicted and the ground-truth boxes respectively, and $w_{i,j}$ the weight assigned to element $j$ in box $i$, typically a $1.0$ without any normalization or re-weighting.
A limitation of the $L_{1}$ is its inability to account for varying sizes of object boxes, i.e., it fails to account for identical deviations in smaller boxes versus larger ones, leading to a bias towards larger boxes.
This discrepancy is especially critical in DETR-based detectors which do not perform any normalization of prediction and ground-truth with respect to any anchor boxes. Furthermore, this bias is propagated between the teacher and student models in the SSOD framework, leading to more severe performance discrepancy for objects of varying sizes. To address this imbalance, we propose a normalization scheme for the regression loss that considers the dimensions of the bounding boxes, ensuring a more equal treatment of objects irrespective of their size. The proposed weighting function, $w_{i,j}^N$ is defined as follows, aiming to adjust the loss based on the width and height of each ground-truth bounding box.
\[
w_{i,j}^N = \begin{cases} 
\frac{1}{\text{width}_i}, & b_{i,j} \in \{x,\text{width}\} \\
\frac{1}{\text{height}_i}, & b_{i,j} \in \{y,\text{height}\}
\end{cases}
\]

As illustrated in \cref{Sized $L_{1}$ loss}, by implementing this normalization, each bounding box is effectively scaled to a constant scale with equal width and height, ensuring equal importance for smaller boxes as larger ones in the loss calculation. To compensate for the potential change in the scale of the loss post-normalization, we can further adjust the weight term by the average dimensions of the bounding boxes prior to normalization.

This normalization strategy can be applied in two different configurations in the SSOD task: (1) across both supervised and unsupervised branches, or (2) only within the supervised branch. 
Additionally, this loss is equally applicable to the fully-supervised setting, such as Co-DETR. The subsequent section will further elaborate on the experimental results.

\section{Experiments}
\subsection{Datasets and Evaluation Metrics}
Our evaluation with Semi-DETR uses the MS-COCO dataset\cite{lin2014microsoft}, a widely utilized benchmark with 80 object classes under two settings following \cite{xu2021end}: (1) \textbf{COCO-partial}. We randomly sample 1\%, 5\%, and 10\% of the train2017 set, which contains 118k images, as labeled data while using the rest as unlabeled data. Furthermore, five different folds are created, and the average of COCO $mAP$ on the val2017 set containing 5k images is reported. (2) \textbf{COCO-full}. We utilize the whole of train2017 as the labeled set and the unlabeled2017 set containing 123k unlabeled images as the unlabeled set. Similarly, the COCO $mAP$ on the val2017 set is reported as the performance metric.

Additionally, we evaluate our method under a fully-supervised setting with Co-DETR, the current SOTA method for DETR-based detector. We train the model with the train2017 images as the supervised data and report the COCO $mAP$ on the val2017 set. 

For both semi-supervised and fully-supervised tasks, we report the $mAP$ across different object sizes --- small for $area < 32^2$, medium for $32^2 < area < 96^2$, and large for $area > 96^2$, following the size convention set in MS-COCO \cite{lin2014microsoft}.

\subsection{Implementation Details}
Following the configurations used in Semi-DETR \cite{zhang2023semi}, DINO \cite{zhang2022dino} is used as with ResNet-50 \cite{he2016deep} as the backbone. The number of object queries is set to 900, following the setting used in DINO. To ensure a fair comparison, we use the same hyperparameters as Semi-DETR when training under the COCO-partial setting: 120k iterations with 8 GPUs and 5 images per GPU. For COCO-full, we train with 240k iterations with 8 GPUs and 5 images, as opposed to 8 images, per GPU. Besides, we use the same set of hyperparameters as Semi-DETR. Furthermore, we vary the iteration where the assignment is changed; more details will be discussed in the Ablation Study section.
For the fully-supervised task, Co-DETR \cite{zong2023detrs} serves as the baseline. We replace the $L_{1}$ regression loss in the main DETR head with Sized $L_{1}$ loss while maintaining the same losses for the auxiliary heads. We train the model with the same setting and hyperparameters as described in Co-DETR to ensure a fair comparison.

\renewcommand{\arraystretch}{1.5} 
\begingroup
\setlength{\tabcolsep}{10pt} 

\begin{table}[tb]
    \centering
    \caption{$mAP$ comparison of SSOD methods with the proposed Sized $L_{1}$ loss under the COCO-partial setting}
    \begin{tabular}{ C{4.5cm}  |  c  |  c  | c  }
        \toprule
        Method & 1\% & 5\% & 10\% \\ 
        \midrule
        Semi-DETR & \textbf{30.5} & 40.1 & 43.5  \\
        Semi-DETR + Sized $L_{1}$ loss & 28.5 & 40.9 & \textbf{43.9} \\
        Semi-DETR + Sized $L_{1}$ loss (Sup only) & 28.8 & \textbf{41.2} & \textbf{43.9} \\ 
        \bottomrule
    \end{tabular}
    
    \label{tab:coco partial overall}
\end{table}
\endgroup

\begin{table}[tb]
    \centering
    \caption{$mAP$ comparison by size under COCO-partial setting}
    \begin{tabular}{ C{4.5cm} | c  c  c | c  c  c | c  c  c }
        \toprule
        \multirow{2}{*}{Method} & \multicolumn{3}{c|}{$mAP_s$} & \multicolumn{3}{c|}{$mAP_m$} & \multicolumn{3}{c}{$mAP_l$} \\
        \cline{2-10}
        & 1\% & 5\% & 10\%
        & 1\% & 5\% & 10\%
        & 1\% & 5\% & 10\% \\
        \midrule
        
        Semi-DETR & \textbf{15.5} & 23.6 & 25.3 & \textbf{32.4} & 42.4 & 45.8 & \textbf{40.2} & 52.6 & 57.9  \\
        Semi-DETR + Sized $L_{1}$ loss & 14.8 & 23.2 & 26.9 & 31.5 & 42.6 & 46.9 & 38.4 & 53.4 & 58.0 \\
        Semi-DETR + Sized $L_{1}$ loss (Sup only) & 14.4 & \textbf{23.7} & \textbf{28.8} & 30.5 & \textbf{42.7} & \textbf{47.4} & 39.7 & \textbf{53.5} & \textbf{58.4} \\ 
        \bottomrule
    \end{tabular}
    
    \label{tab:coco partial size}
\end{table}

\begingroup
\setlength{\tabcolsep}{4pt} 
\begin{table}[tb]
    \centering
    \caption{$mAP$ comparison of SSOD methods with Sized $L_{1}$ loss under COCO-full setting}
    \begin{tabular}{ C{4.5cm} | c | c | c | c }
        \toprule 
        Method & $mAP_{s}$ & $mAP_{m}$ & $mAP_{l}$ & $mAP$ \\ 
        \midrule
        Semi-DETR & 33.2 & 53.7 & 65.7 & 50.4  \\
        Semi-DETR + Sized $L_{1}$ loss & 33.4 & \textbf{54.0} & 65.6 & 50.7 \\
        Semi-DETR + Sized $L_{1}$ loss (Sup only) & \textbf{33.5} & \textbf{54.0} & \textbf{65.7} & \textbf{50.8} \\
        \bottomrule
    \end{tabular}
    
    \label{tab:coco full}
\end{table}

\begin{table}[tb]
    \centering
    \caption{$mAP$ comparison of Co-DETR with Sized $L_{1}$ loss}
    \begin{tabular}{ c | c | c | c | c }
        \toprule 
         Method & $mAP_{s}$ & $mAP_{m}$ & $mAP_{l}$ & $mAP$ \\ 
        \midrule
        Co-DETR & 35.4 & 55.6 & 66.7 & 52.1 \\
        Co-DETR + Sized $L_{1}$ loss & \textbf{38.2} & \textbf{55.8} & \textbf{66.9} & \textbf{53.8} \\
        \bottomrule
    \end{tabular}
    
    \label{tab:co-detr full}
\end{table}

\begin{table}[t]
    \centering
    \caption{Effects of the number of iterations and the change in assignment from one-to-many to one-to-one}
    \begin{tabular}{ c | c | c }
        \toprule 
        \#iter & Assignment change & 10\% \\ 
        \midrule
        \multirow{2}{*}{120k} & Yes & 43.9  \\
        & No & 43.9 \\ \hline
        \multirow{2}{*}{300k} & Yes & 43.9 \\
        & No & \textbf{44.4} \\ 
        \bottomrule
    \end{tabular}
    
    \label{tab:ablation1}
\end{table}

\endgroup

\subsection{Comparison}
\subsubsection{Semi-Supervised Object Detection}
We utilize the current SOTA DETR-based SSOD method, Semi-DETR, and compare it with and without Sized $L_{1}$ loss on the MS-COCO dataset benchmark. 
According to \cref{tab:coco partial overall}, introducing Sized $L_{1}$ loss leads to improvements in overall $mAP$ by 1.1 from 40.1 to 41.2 and 0.4 from 43.5 to 43.9 under the 5\% and 10\% data settings, respectively. We observe a more significant improvement with $mAP$-small metric with a 3.5 $mAP$ improvement under 10\% data setting, from 25.3 to 28.8; refer to \cref{tab:coco partial size} for $mAP$ comparison across different sizes. In addition to COCO-partial, we verify that the normalization improves under COCO-full benchmark by 0.4 $mAP$ from 50.4 to 50.8 as shown in \cref{tab:coco full}. These experiments show that our method improves upon the baseline, particularly for smaller objects.

\subsubsection{Fully-Supervised Object Detection}
As shown \cref{tab:co-detr full}, employing the Sized $L_{1}$ loss in Co-DETR under the fully-supervised setting improves the $mAP$ for small objects by 3.2 from 35.2 to 38.4 while the overall $mAP$ improves by 1.7 from 52.1 to 53.8. This confirms that Sized $L_{1}$ loss improves the performance of object detection under both semi-supervised and fully-supervised settings.

\subsection{Ablation Study}
In this section, we validate the effectiveness of our proposed method under different training configurations: training iterations, assignment change, and effective batch size. Shown in \cref{tab:ablation1}, the best result was obtained with 300k iterations without assignment change with final $mAP$ of 44.4.

\section{Conclusion}
We analyzed the scale bias caused by the $L_{1}$ regression loss in DETR-based detectors and proposed the normalized Sized $L_{1}$ loss to use in conjunction with the existing $IoU$ loss. This normalizes the boxes with respect to their width and height, effectively causing equal-sized boxes to be used in the $L_{1}$ loss, ensuring equal contribution of smaller boxes as that of larger ones to the loss calculation. Our experiments showed the effectiveness of Sized $L_{1}$ loss under semi-supervised scenario using Semi-DETR on different data splits as well as fully-supervised scenario using Co-DETR.

\subsubsection{Acknowledgements} Please place your acknowledgments at
the end of the paper, preceded by an unnumbered run-in heading (i.e.
3rd-level heading).

%
%
%
%




\bibliographystyle{splncs04}
\bibliography{main}
\end{document}